\pdfoutput=1

\documentclass{article} 
\usepackage{iclr2026_conference,times}

\iclrfinalcopy

\usepackage{amsmath,amsfonts,bm}

\def\eqref#1{equation~\ref{#1}}

\def\1{\bm{1}}

\DeclareMathAlphabet{\mathsfit}{\encodingdefault}{\sfdefault}{m}{sl}
\SetMathAlphabet{\mathsfit}{bold}{\encodingdefault}{\sfdefault}{bx}{n}

\usepackage{graphicx}
\usepackage{float}          
\usepackage{booktabs}
\usepackage{multirow}
\usepackage{array}
\usepackage[table]{xcolor}  
\usepackage{url}
\usepackage{hyperref}       

\definecolor{stage}{RGB}{30,90,160}
\definecolor{compute}{RGB}{0,120,0}
\definecolor{loss}{RGB}{160,60,0}
\definecolor{control}{RGB}{120,0,120}
\definecolor{comment}{RGB}{120,120,120}

\title{Think Deep, Speak Once: ReLIT, A Recursive Latent Implicit Transformer Framework}

\author{Abhishek Panwar, Maheep Singh, Saksham Bansal$^{*}$ \\
Indian Institute of Technology Roorkee, India \\
\texttt{\{abhishek\_p1, maheep\_s, saksham\_b\}@me.iitr.ac.in}
}

\begin{document}
\maketitle
\renewcommand{\thefootnote}{\fnsymbol{footnote}}
\footnotetext[1]{%
    \makebox[\dimexpr\textwidth-1.8em\relax][s]{%
        \fontsize{8pt}{9pt}\selectfont
        Equal contribution \hfill Code: \url{https://github.com/mdgspace/Think-Deep-Speak-Once-ReLIT}}%
}
\renewcommand{\thefootnote}{\arabic{footnote}}
\begin{abstract}

Chain-of-Thought (CoT) prompting has become the dominant paradigm for eliciting reasoning in Large Language Models (LLMs), yet it creates substantial computational overhead by forcing models to externalize intermediate reasoning steps as discrete tokens. Recent latent reasoning approaches attempt to internalize this process within continuous hidden states. One of the latest advancements in the field of latent reasoning, Tiny Recursive Models (TRMs) excel at symbolic reasoning but struggle to preserve semantic coherence in natural language settings. To bridge this gap, we introduce \textbf{ReLIT} (\textbf{Re}cursive \textbf{L}atent \textbf{I}mplicit \textbf{T}ransformer), a hybrid framework that grounds deep recursive reasoning within the rich semantic representations of a foundational model. ReLIT augments a frozen LLM backbone (TinyLlama-1.1B) with a lightweight, trainable recursive block that iteratively refines its latent thinking ($z$) before committing to a final output, structurally solving linguistic intuition from algorithmic processing and enabling ``deep thinking'' via gradient-isolated recurrent loops without the latency of explicit token generation. Empirically, ReLIT achieves high parameter efficiency on the GLoRE logical reasoning benchmark, matching or outperforming significantly larger models on challenging tasks such as ProofWriter and RuleTaker despite minimal supervision. These results demonstrate that reasoning capability can be scaled efficiently through recurrent depth rather than parameter width, offering a principled framework for semantically grounded implicit reasoning.

\end{abstract}

\section{Introduction}
Large Language Models (LLMs) have demonstrated remarkable capabilities in complex problem-solving. The dominant paradigm driving this success is \textbf{Chain-of-Thought (CoT)} reasoning \citep{wei2022chain}, where models are prompted to explicitly generate intermediate reasoning steps in natural language before producing a final answer. By verbalizing the thought process, CoT decomposes complex problems into manageable sequential steps.
However, this reasoning methodology comes with significant efficiency drawbacks. CoT incurs substantial computational costs due to the generation of long, verbose reasoning chains, which increase latency and memory consumption. Moreover, while autoregressive natural language generation aligns intuitively with human communication, recent research suggests it may not be the most efficient medium for reasoning. \citep{hao2024training}. Generating discrete tokens for every atomic reasoning step forces the model to collapse high-dimensional ``thoughts'' into low-bandwidth text, potentially losing information and wasting compute on syntactic overhead.
To address this, the field is shifting toward \textbf{Latent Reasoning} \citep{zhu2025surveylatentreasoning}. This paradigm explores performing reasoning steps within continuous hidden representations rather than discrete token space. By internalizing the ``chain of thought'' into a recurrent or deep latent process, models can theoretically reason with greater expressiveness and efficiency.
Despite this promise, a gap remains in how to effectively architect these latent reasoners for natural language. Training recursive models from scratch on language is difficult due to the requirement of large amounts of computational resources. To bridge this gap between robust language understanding and efficient recursive logic, we introduce \textbf{ReLIT}, a framework that combines deep recursive reasoning with the semantic intuition of a foundation model.
\begin{figure}[H]
    \centering
    \includegraphics[width=0.9\textwidth,  clip]{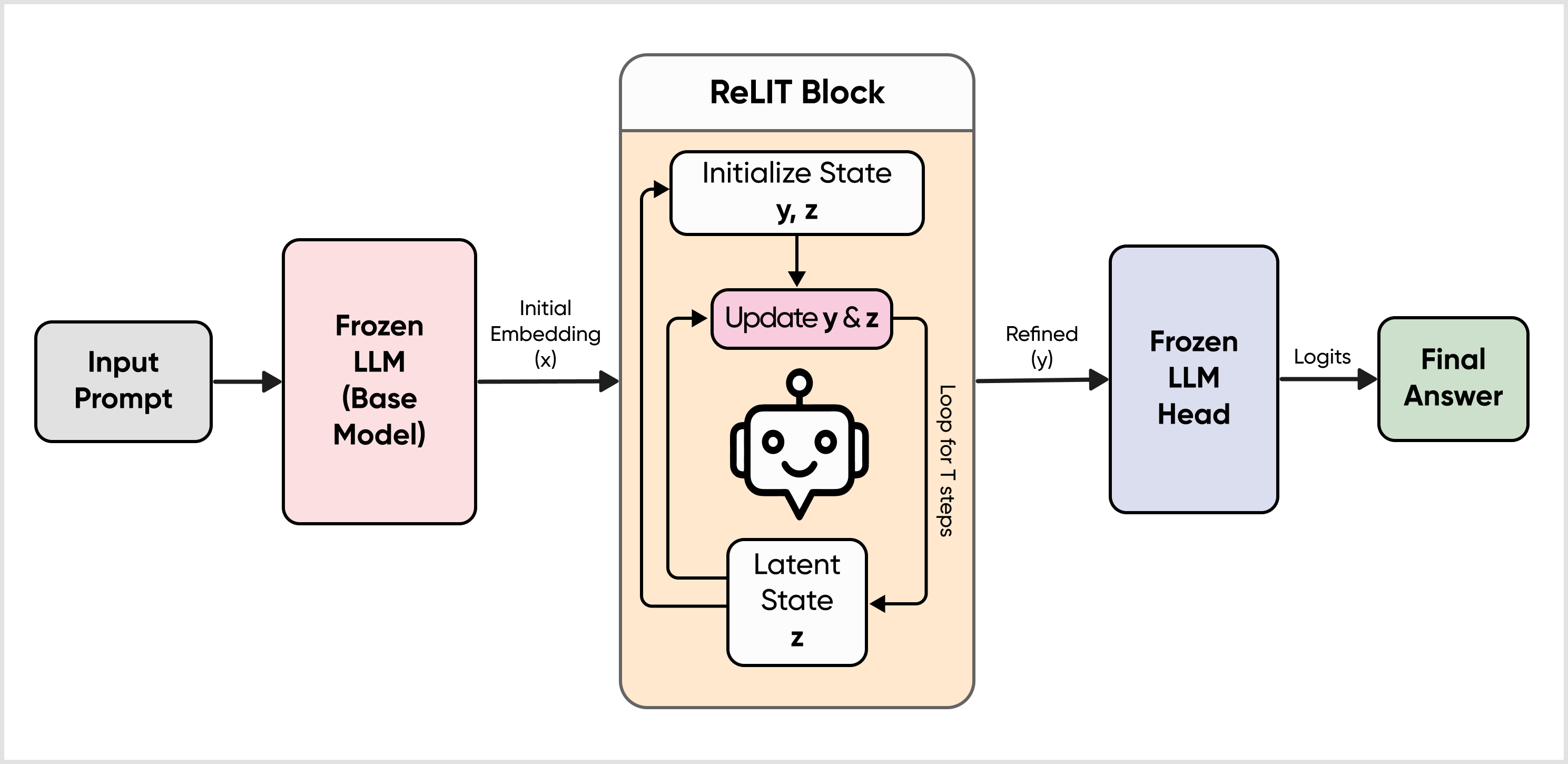}
    \vspace{-2mm} 
    \caption{Inference Pipeline of the ReLIT Framework. The input prompt is processed by a frozen LLM to generate initial embeddings, which are iteratively refined within the recurrent ReLIT Block.}
    \label{fig:inference_pipeline}
\end{figure}
\subsection{Related Work}
Chain-of-Thought prompting for reasoning \citep{wei2022chain} is not only expensive but also single missteps in the intermediary ``logic'' can break the entire reasoning chain. The community has thus come up with many approaches to both increase and make more efficient the reasoning capabilities of an LLM. Such works include Implicit CoT \citep{deng2023implicit} where intermediate steps are removed gradually and the model is finetuned so as to internalize the CoT process. Further works on the concept of internalization and Latent reasoning include COCONUT (Chain of Continuous Thought) \citep{hao2024training} which challenges the inefficiency of human language as a reasoning medium and uses a latent space to reason by feeding the last hidden state as the input embedding for the next token; Dualformer \citep{su2024dualformer} which integrates fast and slow reasoning modes via training on randomized reasoning traces. Parallel works have developed on the idea that recurrent depth is the key to algorithmic reasoning. Work on Universal Transformer \citep{dehghani2019universaltransformers} increased theoretical capabilities via parallel-in-time recurrent self-attentive sequence models. AlgoFormer \citep{gao2024algoformer} increased the algorithmic capabilities of transformers through the combination of a pre-transformer, a looped transformer and a post transformer.
\citet{geiping2025scaling} validated empirically the improved reasoning performance and efficiency that recurrent depth provides over standard Chain-of-Thought reasoning.
\paragraph{Tiny Recursive Models, Hierarchical Reasoning Model and the Semantic Gap}\mbox{}

Hierarchical Reasoning Model \citep{wang2025hierarchical} and Tiny Recursive Model \citep{jolicoeurmartineau2025less} demonstrate the ability of tiny networks on reasoning tasks that LLMs find difficult. Moreover the TRM architecture displays exemplary performance through the use of recursive reasoning. However, these models lack the ability to understand natural language and hence their domain is limited. Our work utilizes ReLIT which draws inspiration from TRM but resolves the semantic limitation with the use of a frozen pre-trained foundational model.

\subsection{The ReLIT Framework: An Introduction}

 \textbf{ReLIT (Recursive Latent Implicit Thinking)} is a hybrid architecture that augments a frozen foundation model with a trainable Recursive Reasoning Block.

Instead of treating reasoning as a single monolithic process, ReLIT structurally decouples it into three distinct, interacting vector states that evolve over time:

\begin{itemize}
    \item \textbf{$x$ (The Semantic Anchor):} A static, high-dimensional intuition extracted from the frozen LLM's last layer. It ensures the model never loses the semantic context of the prompt, no matter how deep the reasoning goes.

    \item \textbf{$y$ (The Answer State):} A dynamic vector representing the model's current hypothesis. It starts as a rough guess and is refined at every step, converging toward the solution.

    \item \textbf{$z$ (The Latent Scratchpad):} A dedicated ``hidden thought'' vector. Unlike the answer state, $z$ is free to explore abstract, high-dimensional logic paths without needing to map to a valid output, acting as a purely internal working memory.
\end{itemize}

By formalizing reasoning as the iterative interaction of these three vectors,where $x$ grounds the thought, $z$ processes the logic, and $y$ captures the result; ReLIT enables deep, multi-step deduction on natural language tasks without the computational overhead of explicit token generation.

\section{Our Methodology}

\begin{figure}[t]
    \centering
    \makebox[\textwidth][c]{
        \includegraphics[width=1.15\textwidth, clip]{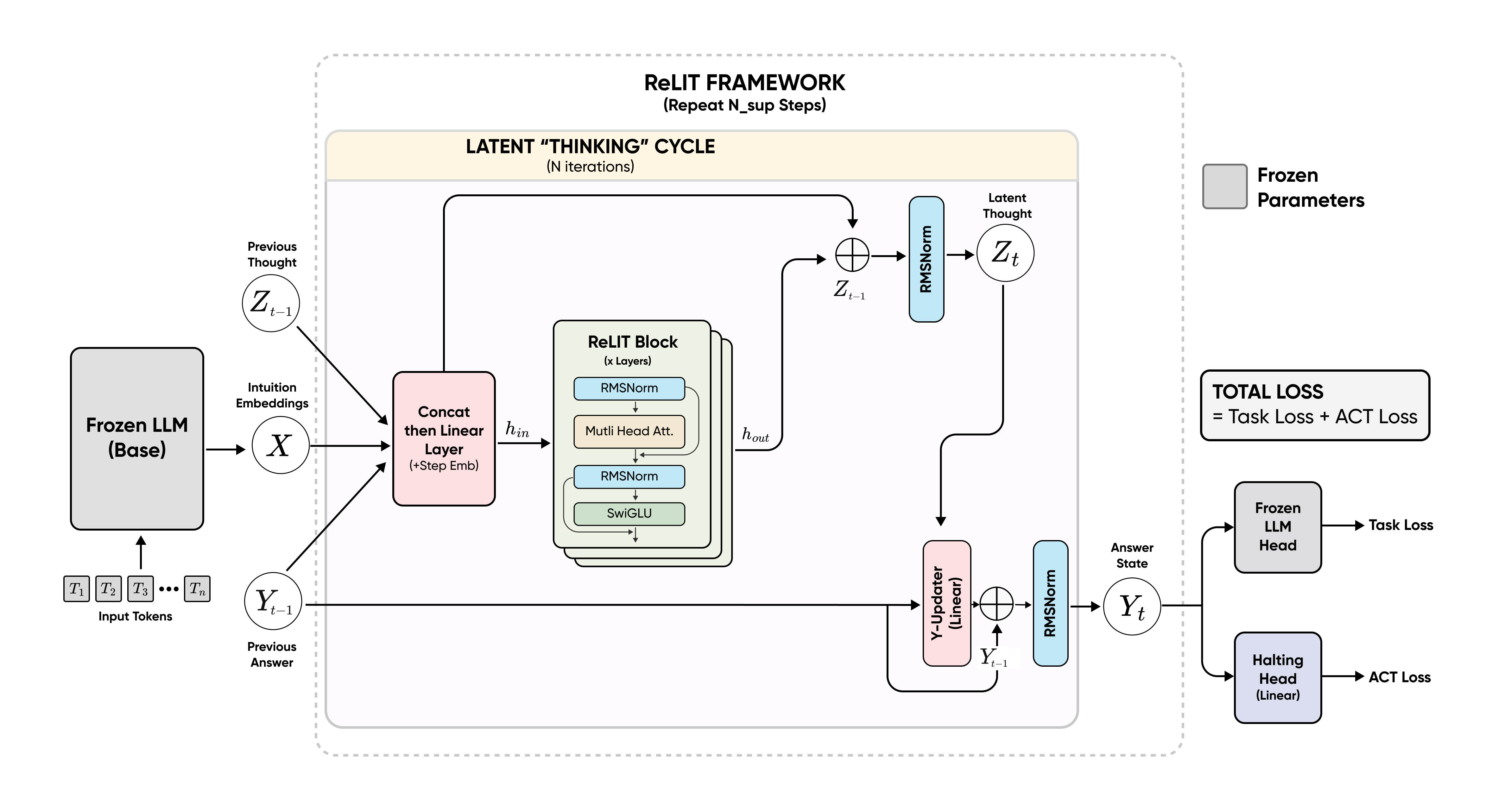}
    }
    \vspace{-3mm}
    \caption{ \textbf{The ReLIT Pipeline.} An overview of the proposed architecture featuring a tripartite structure: (1) Semantic Grounding via a frozen LLM backbone, (2) Recursive Refinement within the latent ReLIT block, and (3) Hypothesis Decoding via the frozen head.}
    \label{fig:main_arch}
\end{figure}

We propose \textbf{ReLIT Framework} as shown in fig.\ref{fig:main_arch} , a tripartite architecture designed to decouple linguistic intuition from logical reasoning. The model follows a ``sandwich'' structure, effectively situating a high-frequency \textit{(Recursive Latent Implicit Thinking)} (ReLIT) block between the frozen layers of a pre-trained Large Language Model (LLM).

\subsection{The Backbone: Semantic Grounding}
The foundation of our architecture utilizes a frozen \textbf{TinyLlama-1.1B} \cite{zhang2024tinyllama} backbone.The primary role of the backbone is to provide \textbf{Semantic Grounding} \cite{lyre2024understandingaisemanticgrounding}. By passing the input through the pre-trained transformer once, we extract the initial latent representation $x \in \mathbb{R}^{B \times L \times D}$, where $D$ is the hidden dimension. This ensures the ReLIT Block inherits a rich, high-dimensional understanding of language, allowing it to focus exclusively on logic and state-refinement.

\subsection{ ReLIT Block}
ReLIT architecture is inspired from Tiny Recursive Models(TRMs) with novel ammendments.
The internal architecture of ReLIT is built upon a high-performance transformer-inspired block, utilizing \textbf{RMSNorm} \cite{zhang2019rootmeansquarelayer} for pre-normalization and a \textbf{SwiGLU} \cite{shazeer2020gluvariantsimprovetransformer} based feed-forward network to ensure the cell possesses the expressive capacity of much larger models while remaining computationally lean. Rather than following the TRM approach of direct state replacement, our implementation introduces a \textbf{Residual State Refinement} strategy as a core structural novelty. In this framework, the model does not attempt to regenerate the thought vector $\mathbf{z}$ and answer state $\mathbf{y}$ from scratch at each step. Instead, it computes a logical \textit{delta}, or refinement, which is added back to the existing states:
\begin{equation}
    {z}_{n+1} = \text{RMSNorm}({z}_n + \text{ReLIT}(\text{Linear}([{x}, {y}_n, {z}_n])))
\end{equation}
By framing the update as $\mathbf{z} + \Delta\mathbf{z}$, the ReLIT acts as a differential reasoning engine. This allows the model to incrementally ``sculpt'' the final answer over $N_{latent}$ iterations and $T$ deep recursions, preserving the foundational linguistic context ${x}$ provided by the backbone while iteratively stripping away logical inconsistencies. This residual dependency ensures smoother gradient flow during the deep supervision process and prevents the vanishing of semantic information during extended ``thinking'' cycles.
\subsection{The Decoder: From Latents to Tokens}
The final component is the Projection Head, which decrypts the evolved latent state ${y}$ into the vocabulary space using a strategy tailored to the dataset's reasoning requirements. By employing a Frozen Base Head, we enforce Semantic Consistency, anchoring the ReLIT's reasoning to the original LLM's semantic map to prevent latent drift. Alternatively, a Trainable Decoder allows for Task-Specific Specialization, granting the model the flexibility to re-map its internal logic to specialized domain tokens however, it requires higher compute and tokens than frozen one. This modular design provides a choice between the regularizing effect of a frozen head for general language and the adaptive precision of a trained head for complex, logic-heavy tasks.

Furthermore, we deliberately employ a Decoder-only backbone (TinyLlama-1.1B) instead of a bidirectional Encoder to extract the semantic intuition . While Encoders often yield superior embeddings for discriminative tasks, a causal Decoder architecture ensures the framework remains forward-compatible with multi-token generation. This choice aligns with our broader research objective to establish ReLIT as a foundation for \textbf{Latent Chain-of-Thought} reasoning, enabling the recursive ``thinking'' phase to be seamlessly integrated into sequential autoregressive generation in future iterations.
\section{Mathematical Formulation}

We formulate ReLIT as a discrete dynamical system that iteratively refines a latent hypothesis. Unlike standard Transformers that map input to output in a single pass, our framework introduces a \textit{time} dimension to the thinking process, allowing the model to revisit and improve its internal state before committing to a prediction.

\subsection{The Latent State Triad}

We define the system state at any recursive step $t$ via three distinct vector fields, where $L$ denotes the sequence length and $D$ represents the hidden dimension.

\begin{enumerate}
    \item \textbf{The Semantic Intuition ($x$):} A static semantic anchor derived from the frozen backbone ($\mathrm{LLM}_{\text{frozen}}$). It remains constant to prevent semantic drift:
    \begin{equation}
        x = \mathrm{LLM}_{\text{frozen}}(input), \qquad x \in \mathbb{R}^{L \times D}
    \end{equation}

    \item \textbf{The Latent Reasoning State ($z_t$):} A ``scratchpad'' memory vector capturing the intermediate chain of thought. It is initialized from a learnable Gaussian distribution:
    \begin{equation}
        z_0 \sim \mathcal{N}(0, \sigma^2 I), \qquad z_t \in \mathbb{R}^{L \times D}
    \end{equation}

    \item \textbf{The Answer State ($y_t$):} The evolving representation of the final output, initialized as a copy of the semantic intuition and refined incrementally:
    \begin{equation}
        y_0 = x, \qquad y_t \in \mathbb{R}^{L \times D}
    \end{equation}
\end{enumerate}

\subsection{Recursive Dynamics and Recall-Then-Learn}

We model reasoning as a discrete dynamical system governed by a transition function $F_\theta$, evolving a latent state $(y_t, z_t)$ conditioned on the static anchor $x$. To enable deep recurrence without the memory overhead of Backpropagation Through Time (BPTT)\cite{bird2025backpropagationtimenetworkslongterm}, we utilize a \textbf{Recall-Then-Learn} protocol. This approximations a stable fixed point via a gradient-free warm-up, ensuring a minimal memory footprint.

\subsubsection{Recursive Transition Function}
Let $h_t$ denote the fused hidden representation at step $t$. The state update $(y_{t+1}, z_{t+1}) = F_\theta(y_t, z_t; x)$ is defined by the following transformation:
\begin{align}
    h_t &= W_c [x \, ; \, y_t \, ; \, z_t] + e_t \\
    z_{t+1} &= \mathrm{Norm}_z \big( z_t + \mathcal{R}_\theta(h_t) \big) \\
    y_{t+1} &= \mathrm{Norm}_y \big( y_t + W_y [y_t \, ; \, z_{t+1}] \big)
\end{align}
where $[-;-;-]$ denotes concatenation, $W_c$ and $W_y$ are learned linear projections, $e_t$ is a step-dependent embedding, and $\mathcal{R}_\theta$ is the recursive Transformer block.

\subsubsection{Recall-Then-Learn Protocol}
We approximate the logical fixed point by decoupling the training depth from the inference depth through two distinct phases:

\textbf{Phase I: Recall (Warm-up).} Updates are performed with gradients disabled to allow the system to settle into a stable reasoning path:
\begin{equation}
    (y_{t+1}, z_{t+1}) = F_\theta(y_t, z_t; x), \qquad t = 0, \dots, T-2
\end{equation}

\textbf{Phase II: Learn (Optimization).} A single gradient-enabled update is performed on the matured state:
\begin{equation}
    (y_T, z_T) = F_\theta\big(\mathrm{stopgrad}(y_{T-1}, z_{T-1}); x\big)
\end{equation}

By detaching the states between supervision steps (BPTT) \cite{bird2025backpropagationtimenetworkslongterm}, the framework enables deep, multi-step deliberation while maintaining a manageable memory footprint.

\subsubsection{Projection and Inference}

The final refined answer state is projected back into the vocabulary space using the frozen output projection of the base LLM ($W_{\text{lm}}$):
\begin{equation}
    p(\cdot \mid s) = \mathrm{Softmax}\big( W_{\text{lm}} \, y_T \big)
\end{equation}

During training, we apply \textbf{Deep Supervision \cite{lee2014deeplysupervisednets}} by computing cross-entropy loss on $y_t$ at each macro-step. This stabilizes training by encouraging intermediate states to remain aligned with the target output throughout the recursion.

\subsection{Recursive Latent Refinement with Adaptive Halting}
The predefined states of y and z are iteratively refined through a series of N supervision steps, each containing T recursive passes. We utilize residual updating and Layer Normalization to ensure training stability. \newline
To optimize the computation budget we employ Adaptive Computation Time (ACT)\cite{graves2017adaptivecomputationtimerecurrent} mechanism where a Halt Head predicts a confidence-based halting probability, allowing the model to dynamically mask and scale the loss for samples that have reached logical convergence.
\section{Training Methodology}
To better illustrate the training methodology, this subsection will delve in detail into the process. Please refer to the pseudocode provided in the appendix.

We start by sending the input prompt to the frozen base model (TinyLlama) and extract the hidden layer ${x}$ from the last layer. Then we initialize ${y}$ (current answer) and ${z}$ (latent internal reasoning).
The input for the ReLIT Block ${h_{\text{in}}}$ is created by concatenating ${x}$,${y_{t-1}}$,${z_{t-1}}$; passing them through a linear layer and adding step embeddings. The ReLIT Block refers to a recursive Transformer Block containing a RMS Normalization layer, MutliHead Attention layer, another RMS Normalization layer and a SwiGLU Network. The output of the ReLIT Block ${h_{out}}$ is used for the residual updating of ${z_t}$. The underlying process was inspired from Tiny Recursive Model(TRMs)'s training process. We perform this latent reasoning step ${N_{latent}}$ times per update of ${y}$ (the current answer). This update of ${y_{t}}$ can be referred to as latent recursion. We perform latent recursion ${T-1}$ times without gradient and then perform a single gradient-enabled update on the matured state. This is known as Truncated BPTT (Backpropagation Through Time). We then speak i.e. project back to tokens using the frozen LLM Head.

For every Epoch we perform the aforementioned steps ${N_{supervision}}$ times and calculate the task loss and the ACT loss. The task loss is the Cross Entropy loss between the logits produced and logits expected. The ACT loss is the Binary Cross Entropy loss, which predicts if we have reached the correct output and can halt. It allows us to scale the loss for samples that have reached logical convergence.

\section{Results and Discussion}
\subsection{Evaluation on GLoRE}

To validate the logical depth and recursive reasoning capabilities of our framework, we conduct extensive evaluations on the \textbf{General Logic Reasoning Evaluation (GLoRE) \cite{liu2025gloreevaluatinglogicalreasoning}} benchmark. We select five diverse datasets: \textit{ProofWriter} \cite{tafjord-etal-2021-proofwriter}, \textit{RuleTaker} \cite{clark-etal-2020-transformers}, \textit{HELP} \cite{help2021}, \textit{NaN-NLI} \cite{truong2022negationbenchmarknannlitest}, and \textit{TaxiNLI} \cite{joshi-etal-2020-taxinli}---to rigorously assess performance across propositional logic, counterfactual reasoning, and multi-hop path planning tasks.

While these benchmarks are traditionally evaluated under few-shot prompting paradigms, the parameter efficiency of ReLIT enables direct supervision without the prohibitive cost of full-model fine-tuning. Accordingly, we freeze the base foundation model and train only the ReLIT block on the designated training splits, reporting accuracy on held-out evaluation sets. We consider a wide range of data regimes, from extremely low-resource settings (e.g., NaN-NLI with 200 samples) to larger logical corpora (e.g., ProofWriter with 5{,}000 samples), demonstrating the adaptability of ReLIT across supervision scales. Detailed dataset statistics and training hyperparameters are provided in Appendix~A.

\begin{table*}[ht]
\centering
\small
\addtolength{\tabcolsep}{4pt}
\begin{tabular}{l ccc cc c}
\toprule
\textbf{Model} & \multicolumn{3}{c}{\textbf{NLI}} & \multicolumn{2}{c}{\textbf{TF}} & \textbf{Avg} \\
\cmidrule(lr){2-4} \cmidrule(lr){5-6}
& \textbf{HL} & \textbf{TN} & \textbf{NN} & \textbf{RT} & \textbf{PW} & \\
\midrule
RoBERTa & 39.47 & 49.91 & \textbf{90.02} & 53.50 & 55.92 & 57.76 \\
\midrule
LLaMA & 32.26 & 41.91 & 47.29 & 48.89 & 53.78 & 44.83 \\
Falcon & 28.49 & 44.66 & 53.31 & 56.11 & 53.33 & 47.18 \\
Mixtral-8x7B & 33.27 & 40.86 & 50.13 & 46.84 & 44.80 & 43.18 \\
\midrule
ChatGPT & 42.13 & 57.30 & 56.59 & 54.74 & 53.95 & 52.94 \\
GPT-4 & 46.01 & 60.08 & 76.74 & 60.19 & 59.66 & 60.54 \\
\midrule
o1 mini & \textbf{63.69} & 81.41 & 71.55 & 75.90 & 75.33 & 73.58 \\
DeepSeek R1 & 62.05 & 75.74 & 72.58 & 75.29 & 80.51 & 73.23 \\
QwQ-32B & 61.53 & \textbf{81.96} & 75.59 & 78.10 & 82.40 & \textbf{75.92} \\
\midrule
\rowcolor{gray!10}
\textbf{ReLIT (Ours)} & 53.43 & 56.30 & 55.20 & \textbf{97.60} & \textbf{98.60} & 72.23 \\
\bottomrule
\end{tabular}

\vspace{0.8em} 
\caption{\textbf{Performance on select GLoRE benchmark tasks.} Tasks are grouped by category: Natural Language Inference (NLI) and Theorem Finding / Logic (TF). \textit{HL}: HELP, \textit{TN}: TaxiNLI, \textit{NN}: NaN-NLI, \textit{RT}: RuleTaker, \textit{PW}: ProofWriter. All results are accuracy (\%). Best results are in \textbf{bold}.Results for LLMs have been taken from GLoRE \cite{liu2025gloreevaluatinglogicalreasoning} }
\label{tab:glore_results}
\end{table*}
\subsection{Dataset Explainations}

\subsubsection{ProofWriter}
\begin{figure}[t!] 
    \centering
    \makebox[\textwidth][c]{
        \includegraphics[width=0.6\textwidth, clip]{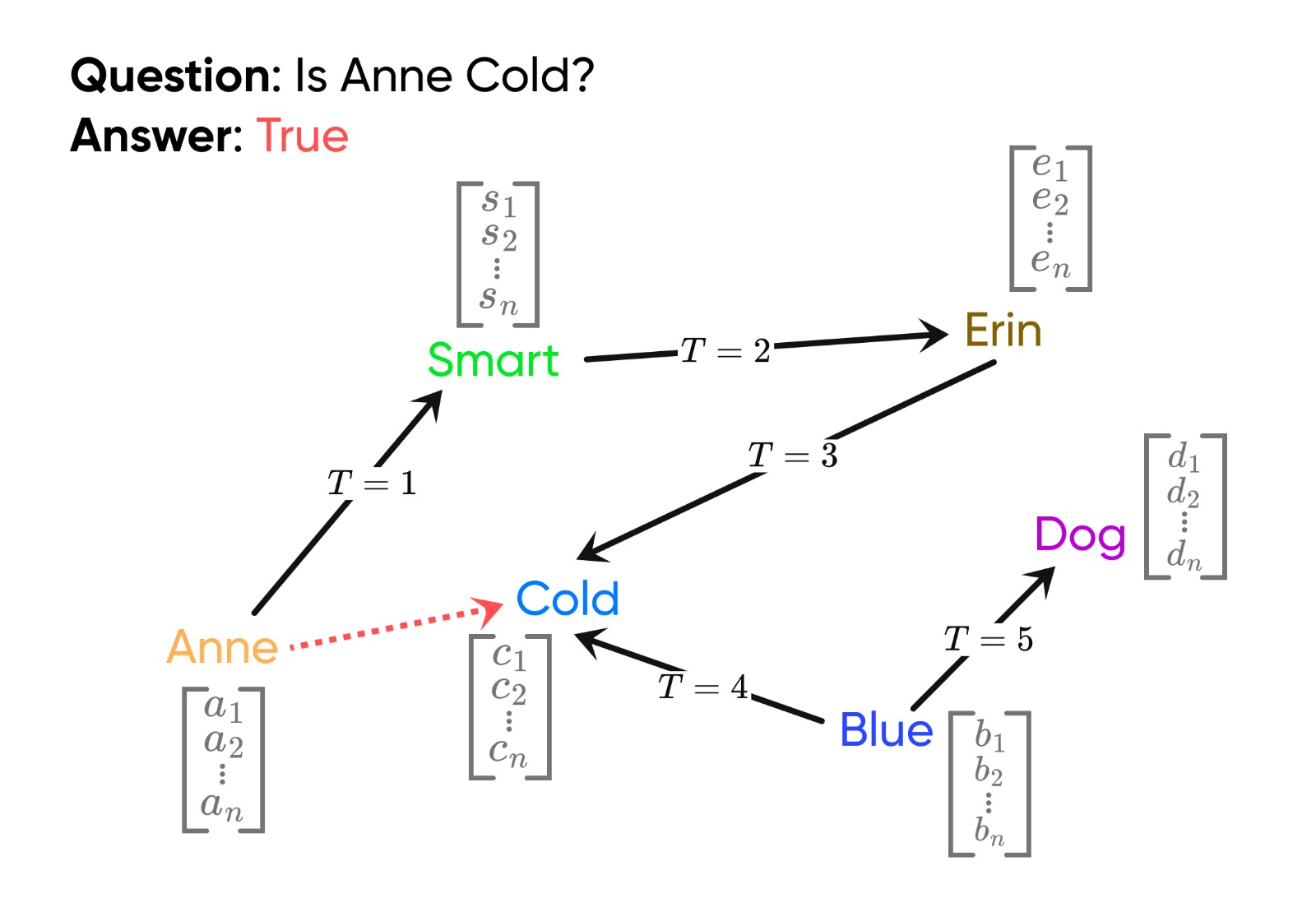}
    }
    \vspace{-2mm} 
    \caption{\textbf{Latent Reasoning Trajectory on ProofWriter.} This visualization depicts the iterative evolution of the state vector within the continuous latent space. Starting from the initial grounding (\textit{Anne}), the model continuously updates its output vector $\vec{y}$ across $T$ steps, transitioning through intermediate logical predicates (\textit{Smart} $\rightarrow$ \textit{Erin} $\rightarrow$ \textit{Cold}). The trajectory demonstrates how the recurrent ReLIT block refines high-dimensional ``hidden thoughts'' to converge onto the final logical state (\textit{Cold}) without generating intermediate tokens.}
    \label{fig:proofwriter_intuition}
\end{figure}

ProofWriter evaluates deductive logic by requiring the model to chain multiple implications (e.g., $A \to B \to C$). We trained on 5,000 samples and tested on 3,000. The model exhibits a clear ``thinking'' phase, halting at an average depth of $3.6$ steps as it moves from initial uncertainty to a stable logical fixed point. This haulting value signifies that model's answer are consistent after 3-4 supervision loops. This process works by mathematically shifting the entity vector (e.g., \textbf{Bob}) through semantic regions (e.g., \textbf{Feline} $\to$ \textbf{Carnivore}) via residual updates until the logic is resolved (Intution example is shown in fig \ref{fig:proofwriter_intuition}). Essentially, the ReLIT treats reasoning as a geometric navigation task, where the final stable vector position directly maps to the correct logical verdict. A remarkable accuracy of {98.6\%} as given in \textbf{Table1} shows the latent and Implicit thinking of \textbf{ReLIT} framework which is even higher than top LLM performers on the particular data.

\subsubsection{RuleTaker}
The RuleTaker dataset similarly evaluates the model's capacity for rule-based inference, often involving larger rule sets and more complex distractors. We trained it on 5000 samples and evaluated on 3000 samples and observed that model follows the same geometric intuition established in the ProofWriter task, utilizing the recursive loops to navigate through the logical constraints of the prompt. For this dataset, the average halting depth increases to \textbf{5.2 steps}. This slightly higher value reflects the increased complexity of the rules, requiring the ReLIT-Block to perform additional residual updates to reach a stable logical fixed point. Despite the increased depth, the underlying process remains consistent as can be noticed in fig \ref{fig:supervision_steps}: the model incrementally shifts the latent states until the entity-property relationships are resolved. The fact that the halting depth naturally scales with task complexity proves that the ReLIT is dynamically adapting its ``thinking time'' to the difficulty of the rule set. The accuracy of {97.6\%} again prooves our architecture capability to reason on logical dataset without Chain-of-Thought(CoT).
\subsubsection{NaN-NLI}
The NaN-NLI dataset presents a unique challenge due to extreme data scarcity, with only 200 samples available for training and 50 for evaluation. Despite these severe computational and data constraints, the model achieves good accuracy comparable to models such as ChatGPT, demonstrating the robust semantic grounding provided by the frozen LLM backbone.

\begin{figure}[t!]
    \centering
    \begin{minipage}{1\textwidth}
        \centering
        \includegraphics[width=\linewidth, clip]{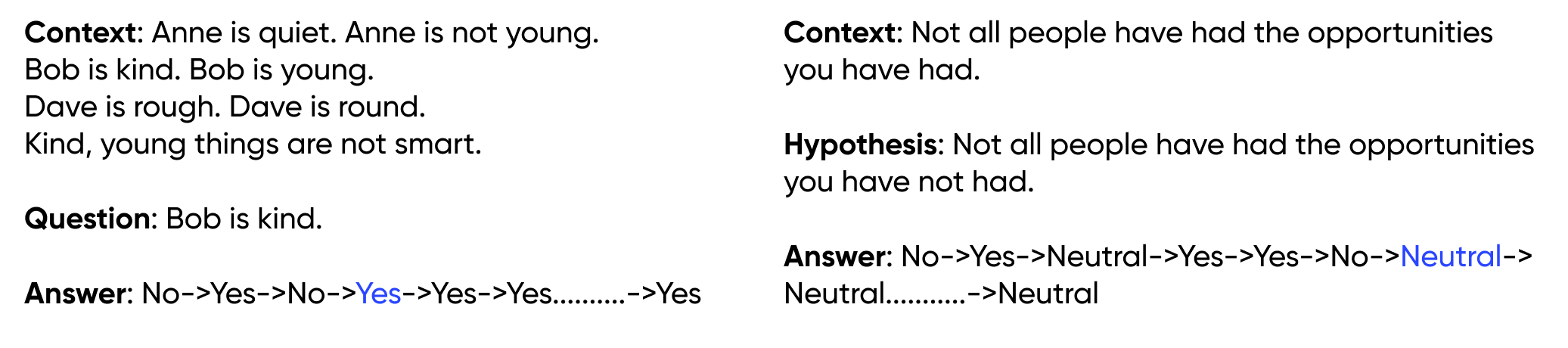}
    \end{minipage}

    \vspace{2mm}
    \caption{\textbf{Latent Reasoning Trajectories and Adaptive Halting.} Visualization of $N$ supervision steps during ReLIT inference on \textit{RuleTaker} (left) and \textit{NaN-NLI} (right). The trajectories illustrate the iterative refinement of the state vector $\bm{v}$. The \textcolor{blue}{blue highlights} denote the dynamically identified \textbf{halting steps}, where the latent state achieves numerical stability, indicating logical convergence before final answer projection.}
    \label{fig:supervision_steps}
\end{figure}

Due to the limited training signal,we get \textbf{55.2\%} accuracy which is still comparable to \textbf{ChatGPT's} accuracy. The model exhibits a higher average halting depth of \textbf{7.8 steps} which can be noticed in fig. \ref{fig:supervision_steps}. This increased depth illustrates the consequences of low-data regimes: the ReLIT must perform more internal recursions to stabilize its logical state and resolve the linguistic nuances of the inference task. However, the critical finding is that the outputs remain consistent after this stabilization phase. The high confidence in the final supervision loops proves that the model is not hallucinating or making stochastic guesses; rather, it is performing deliberate reasoning to overcome the lack of extensive training data. This result reinforces our conclusion that the ReLIT-backbone coupling is a highly sample-efficient architecture for logical tasks.
\subsubsection{HELP}
On the HELP benchmark, we trained on 800 samples and evaluated on 200, achieving a notable accuracy of \textbf{53.43\%} even beating \textbf{ChatGPT} and \textbf{GPT-4}. While this result is competitive with much larger models, the core significance lies in the model's reliability; with an average halting depth of \textbf{6.8 steps}, the \textbf{ReLIT} framework demonstrates that it is not relying on stochastic guesses or hallucinations. Instead, the stabilization of the latent state across recursive loops ensures that the final answers are the product of consistent logical reasoning. This proves that our \textbf{ReLIT} is succesfully reasoning on logical datasets instead of random guesses.
\subsubsection{Taxi NLI (TN)}
On the Taxi NLI (TN) benchmark, ReLIT demonstrates robust reasoning capabilities
despite limited training due to computational constraints we trained it on 2000 samples and evaluated on 800 samples. While achieving an
accuracy of \textbf{56.30\%}, the model's core significance lies in its
avoidance of stochastic guesses and hallucinations. By stabilizing
the latent state across recursive loops, ReLIT ensures that its outputs are the
product of consistent logical inference rather than random patterns.
This proves the model's ability to maintain logical integrity even in complex
NLI tasks with reduced training overhead.
\begin{figure}[t!]
    \centering
    \begin{minipage}{1\textwidth}
        \centering
        \includegraphics[width=\linewidth, clip]{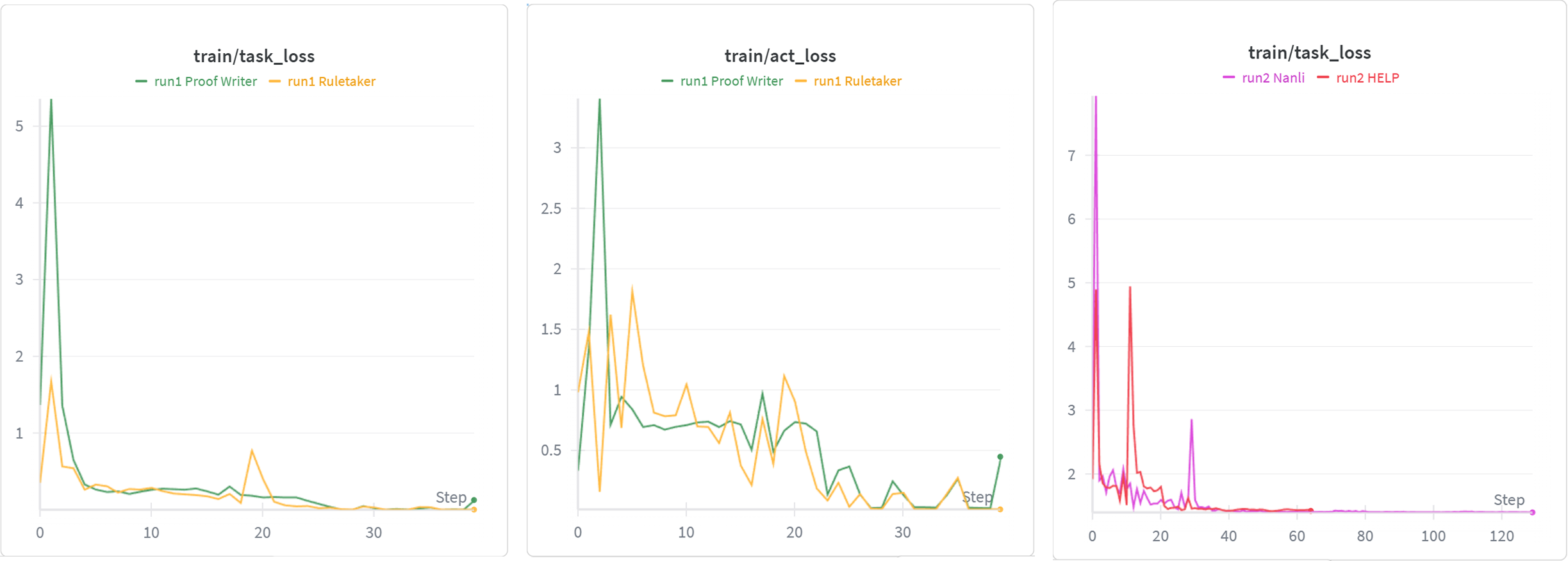}
    \end{minipage}

    \vspace{2mm}
    \caption{\textbf{Training Dynamics and Convergence Profiles.} Visualization of the optimization process across multiple benchmarks. The left and center panels show the \textbf{task loss} and \textbf{ACT loss} for \textit{ProofWriter} and \textit{RuleTaker}, respectively, demonstrating a sharp initial descent followed by stability. The right panel illustrates the \textbf{task loss} for \textit{NaN-NLI} and \textit{HELP}, where the smooth convergence toward a lower bound indicates effective latent state grounding and reliable learning of logical dependencies over training steps.}
    \label{fig:training_convergence}
\end{figure}

\section{Future Work and Conclusion}

\subsection{Future Work: Deep Thinking, Singular Expression}
The current implementation of the \textbf{ReLIT} block demonstrates that recursive latent structures can inherit deep language understanding via implicit reasoning over frozen LLM memory. We envision several high-impact directions for this architecture:

\begin{itemize}
    \item \textbf{Specialized Semantic Encoders:} ReLIT can serve as a sophisticated replacement for BERT \cite{devlin2019bertpretrainingdeepbidirectional}-based models, specifically in tasks like \textbf{FinBERT} \cite{huang2023finbert} for financial sentiment analysis. In these domains, the model must ``think deep but speak once'' i.e. process complex market nuances through multiple latent recursions before outputting a single, high-integrity semantic score.
    \item \textbf{Recursive Classifiers:} Beyond generative tasks, ReLIT can be utilized as a robust classifier for document-level entailment or paragraph-to-paragraph semantic matching, where the recursive loops allow the model to resolve contradictions that standard feed-forward encoders miss.
    \item \textbf{Scalability to Multi-Token Generation:} While our current work focuses on single-token reasoning (thinking deeply before providing a definitive answer), a natural extension is the transition to multi-token autoregressive training. Although this presents significant computational challenges and requires high-density hardware for gradient accumulation across recursive steps, such an approach could allow the ReLIT block to maintain a continuous ``stream of consciousness'' during long-form generation. If computational resources permit, this would test whether the implicit vector math observed in our study scales to the complex dependencies of full-paragraph synthesis.
\end{itemize}

\subsection{Conclusion}
In this paper, we introduced a novel ``sandwich'' architecture that decouples linguistic intuition from logical reasoning. By situating the \textbf{ReLIT} block within the latent space of a frozen LLM, we demonstrate that models can achieve high-performance logical reasoning without the need for explicit Chain-of-Thought (CoT) prompting.Our results across benchmarks like ProofWriter and HELP prove that implicit, latent thinking is both data-efficient and mathematically stable. We conclude that the future of robust AI lies not in simply scaling parameter counts, but in developing specialized recursive engines that can leverage pre-trained semantic memories to solve the underlying geometric structure of logic.

\section*{Acknowledgments}
We thank the members of MDG SPACE, IIT Roorkee for their support and encouragement throughout this work.

\bibliography{iclr2026_conference}
\bibliographystyle{iclr2026_conference}

\appendix
\section{Appendix}

\subsection{Convergence Analysis and the Information Bottleneck}

In this appendix, we analyze the convergence properties of the ReLIT framework. In particular, we study the dependence of the recursive reasoning process on the initial semantic embedding extracted from the frozen backbone. We show that the model's ability to converge to a correct solution is fundamentally bounded by the information preserved by this embedding, resulting in a theoretical \emph{semantic bottleneck}.

\subsection{Fixed-Point Stability and Dependence}

We model ReLIT inference as a discrete dynamical system governed by a transition operator $F_\theta$. Let $s$ denote the input prompt and $y^\ast$ the corresponding ground-truth target. The frozen foundation model $\mathcal{M}$ projects the prompt into a static semantic anchor
\begin{equation}
x = \mathcal{M}(s).
\end{equation}

Let $(y_t, z_t)$ denote the answer and latent reasoning states at recursion step $t$. Under the assumption that the spectral radius of the Jacobian of the transition operator satisfies
\begin{equation}
\rho\!\left(\frac{\partial F_\theta}{\partial (y,z)}\right) < 1,
\end{equation}
the recursive sequence $\{(y_t, z_t)\}_{t=0}^\infty$ converges to a unique fixed point $(y^\dagger, z^\dagger)$ by the Banach Fixed-Point Theorem. The fixed point is defined implicitly as
\begin{equation}
(y^\dagger, z^\dagger) = F_\theta(y^\dagger, z^\dagger; x).
\end{equation}

Since the initial states $y_0$ and $z_0$ are either deterministic functions of $x$ or random noise drawn independently of $s$, the resulting fixed point is effectively a deterministic function of the anchor $x$ alone. We therefore define the asymptotic mapping
\begin{equation}
y^\dagger = \Phi_\theta(x).
\end{equation}

This formulation reveals a critical limitation: regardless of the recursive depth, the reasoning capacity of $F_\theta$ is restricted to transforming and reparameterizing the information already present in $x$. The recursion cannot recover information about $s$ that was discarded during the projection $\mathcal{M}(s)$.

\subsection{The Information Bottleneck}

We formalize this limitation using information-theoretic arguments. Consider the Markov chain induced by the ReLIT architecture:
\begin{equation}
s \;\longrightarrow\; x \;\longrightarrow\; y^\dagger.
\end{equation}

By the \emph{Data Processing Inequality}, any post-processing of $x$ cannot increase its mutual information with the original prompt $s$. Consequently, the mutual information between the predicted output and the input is bounded as
\begin{equation}
I(s; y^\dagger) \leq I(s; x).
\end{equation}

For the model to converge to the correct solution $y^\ast$, the anchor $x$ must act as a sufficient statistic of the prompt for the task. Formally, convergence is achievable if and only if the conditional entropy satisfies
\begin{equation}
H(y^\ast \mid x) \rightarrow 0.
\end{equation}

Equation~(A.7) captures the central bottleneck of the ReLIT framework. If the frozen backbone $\mathcal{M}$ is lossy---discarding semantic nuances required to infer $y^\ast$---then $H(y^\ast \mid x) > 0$. In this regime, the fixed point $y^\dagger$ will converge to a suboptimal solution regardless of the recursive depth, as the reasoning engine operates on an incomplete semantic representation.

\subsection{Relaxation via Trainable Injection}

To mitigate this bottleneck, we consider relaxing the assumption that the semantic anchor $x$ is fixed. By injecting a small set of trainable parameters $\psi$ into the backbone (e.g., via Low-Rank Adaptation), the embedding becomes
\begin{equation}
x_\psi = \mathcal{M}_\psi(s),
\end{equation}
allowing the representation to adapt to the downstream reasoning task.

The optimization objective then implicitly shifts from minimizing only the task loss to increasing the mutual information between the anchor and the target:
\begin{equation}
\max_{\psi} \; I(x_\psi; y^\ast).
\end{equation}

By optimizing $\psi$, the backbone learns to preserve task-relevant semantic features, effectively reducing the conditional entropy in Eq.~(A.7) and ensuring that the fixed point $y^\dagger$ lies within the valid solution manifold.

Alternatively, introducing a cross-attention mechanism that allows the reasoning state to query the prompt representation directly at each recursion step would bypass the bottleneck in Eq.~(A.6), effectively transforming the anchor into a dynamic variable $x_t$.

\subsection{Experiments Hyperparameters}
\begin{table}[t]
\centering
\small
\setlength{\tabcolsep}{4pt}
\begin{tabular}{lccccc}
\hline
\textbf{Dataset} & \textbf{Nsup} & \textbf{Head} & \textbf{Layers} & \textbf{BS} & \textbf{LR} \\ \hline
HELP        & 8  & 16 & 4 & 16  & $8\mathrm{e}{-5}$ \\
TaxiNLI     & 12 & 16 & 4 & 64  & $8\mathrm{e}{-5}$ \\
NaNLI       & 16 & 16 & 4 & 16  & $8\mathrm{e}{-5}$ \\
RuleTaker   & 6  & 16 & 4 & 128 & $1\mathrm{e}{-4}$ \\
ProofWriter & 3  & 8  & 1 & 256 & $2\mathrm{e}{-5}$ \\ \hline
\end{tabular}
\caption{Hyperparameters for Different Datasets}
\label{tab:hyperparams}
\end{table}
\begin{figure}[t] 
    \centering
    \includegraphics[width=0.84\linewidth, trim=2 2 2 2, clip]{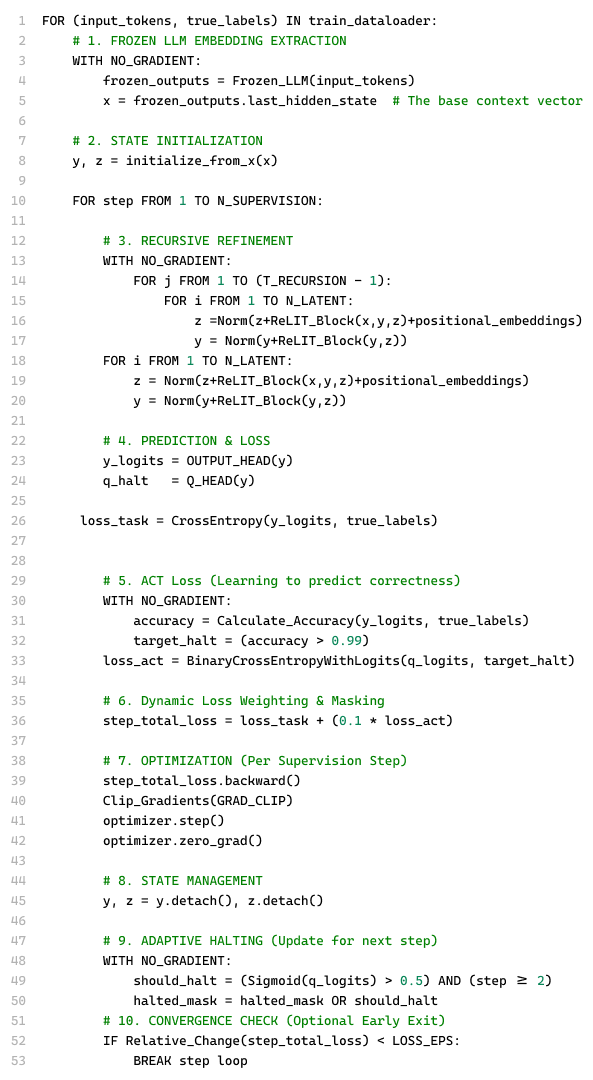}
    \caption{\textbf{Pseudocode of the ReLIT Recursive Update Process.} The algorithm details the initialization of the latent scratchpad $z$, the $T$-step recurrent loop, and the final decoding phase.}
    \label{fig:pseudocode}
\end{figure}

\end{document}